\documentclass[10pt,twocolumn]{article}

\usepackage{graphicx} 
\usepackage{amsmath}
\usepackage[version=4]{mhchem}
\usepackage{siunitx}
\usepackage{longtable,tabularx}

\usepackage{setspace}
\usepackage{amsmath}
\usepackage{fixmath}  
\usepackage{array}
\usepackage{hyperref}
\usepackage{xcolor}
\usepackage{soul}
\usepackage{multirow}

\title{Symplectic Integration for Multivariate Dynamic Spline-Based Model of Deformable Linear Objects}
\author{Alaa Khalifa* and Gianluca Palli\\
Department of Electrical, Electronic and Information Engineering,\\ University of Bologna, Italy\\
*Department of Industrial Electronics and Control Engineering, \\Faculty of Electronic Engineering, Menoufia University, Egypt\\
Email: alaa.khalifa@el-eng.menofia.edu.eg }

\date{}

\begin{document}
\maketitle	
	
	\begin{abstract}
		Deformable Linear Objects (DLOs) such as ropes, cables, and surgical sutures have a wide variety of uses in
		automotive engineering, surgery, and electromechanical
		industries. Therefore, modeling of DLOs as well as a computationally efficient way to predict the DLO behavior are of great importance, in particular to enable robotic manipulation of DLOs. 
		The main motivation of this work is to enable efficient
		prediction of the DLO behavior during robotic
		manipulation. In this paper, the DLO is modeled by a multivariate
		dynamic spline, while a symplectic integration method is used to
		solve the model iteratively by interpolating the DLO shape during
		the manipulation process. Comparisons between the symplectic,
		Runge-Kutta and Zhai integrators are reported. The
		presented results show the capabilities of the symplectic integrator
		to overcome other integration methods in predicting the DLO
		behavior. Moreover, the results obtained with different sets of model
		parameters integrated by means of the symplectic method are reported to show how they influence the DLO behavior estimation.
		
	\end{abstract}
	

	\section{Introduction}
	Many of the objects we handle every day are highly deformable and with
	prevalent plastic behavior. However, humans manipulate those objects
	naturally, with high dexterity, and without any particular issue. In facts,
	manipulating these deformable objects has a wide variety of uses in
	domestic facilities and healthcare, such as robotic surgery
	\cite{jayender}, assistive dressing, garment sorting or folding
	clothing \cite{masey}, \cite{ramisa}. Moreover, it is involved in
	manufacturing, aerospace \cite{shah}, automotive, and
	electromechanical industries generally
	\cite{jiang},\cite{hermansson}.
	
	On the contrary, the manipulation of deformable objects is still a
	challenging activity for robots. This is the main reason why many
	assembly procedures involving such deformable objects are still
	performed manually. One of the main reasons why robots have such
	limitations in deformable object manipulation is due to their complex
	behaviors, unpredictable initial configuration, and limited
	capability in measuring their state.
	
	A thoughtful survey can be found in \cite{sanchez} that focuses on
	deformable object manipulation by robots in industrial and domestic
	applications. Actually, the dynamics of deformable objects
	is complex and nonlinear. Therefore, the state estimation of a
	deformable object is challenging, and the forward prediction is
	expensive. Robust and effective methods to manipulate deformable
	objects and predict their behavior remain extremely difficult to build, despite the several
	applications and attempts made by the robotics community
	\cite{sanchez}.
	
	In this work, the numerical integration of Deformable Linear Objects (DLOs),
	such as strings, cables, electric wires, catheters, ropes, and so on,
	is addressed, since efficient solutions to this problem will enable the implementation of robotized solution in many relevant subfields of the large and
	diverse industrial manufacturing. DLO manipulation is fundamental in automotive
	manufacturing, e.g. for wiring harness preparation and electrical cable installation inside the vehicle structure, as well as in medical surgery, e.g.
	in suturing in which a flexible wire in a straight configuration
	needs to go to a knot \cite{moll}, and have a vital role in
	other fields such as architecture and power distribution.
	The relevant work on DLO manipulation can be split into three overlapping categories: modeling,
	simulation, and planning. Various manipulation tasks such as knot
	tying \cite{huang}, rope untangling \cite{tangled}, string insertion
	\cite{wang}, and shape manipulation \cite{rambow} can be executed on
	DLOs.
	
	A model of DLO deformation/flexibility is needed in order to predict its behavior successfully. This model implies that both the
	geometry and the mechanical behavior of the concerned parts can be
	represented accurately. Once the model is properly defined, a
	computationally efficient way is needed to evaluate it over time or to
	solve queries of motion planning. Virtual prototyping is used to
	reduce development costs and to boost consistency. Also, it enables
	the early detection of possible problems. It allows for studying the
	efficiency of assembly. The most distinctive characteristic of DLOs is
	the variations in shape that they undergo following the influence of
	forces and environmental constraints. Such shape variations may be the
	purpose of the planning itself, for example if they are caused by
	obstacles along the path \cite{jimenez}. Overall, DLO modeling is a
	major and complex challenge, with a broad range of
	applications. 
	The PDE model can be established \cite{hu2, hu3, hu4, hu5, hu6}, and the associated structure-preserving method can be developed for the dynamic problems of the DLOs.
	Various research in the literature focused on modeling
	and manipulating DLOs for various purposes and many models and
	strategies were created. There are several various approaches for
	physically modeling DLOs such as Mass-spring \cite{lvphysically},
	Multi-body \cite{servin}, Elastic rod \cite{linn}, Dynamic spline
	\cite{valentini}, Finite element \cite{greco}, and other models. The
	methodology, advantages and disadvantages of each model method are
	discussed in detail in \cite{lv}. The dynamic spline is one of these
	methods which provide a good theoretical basis, continuous model,
	higher authenticity. A geometrically reliable model of DLOs is
	generated and adopted in \cite{theetten} to execute numerical
	simulations on the motion of the object under gravity and during
	environmental interaction. For interaction simulation, a linearized
	spline-based DLO model named quasi dynamic splines is created in
	\cite{quasi}.
	
	Owing to the inherent trade-off between precision and real-time
	capability, it is hard to deal with DLOs due to their complex model,
	resulting then in time-consuming integration processes to predict
	their behavior. To solve numerically the differential equations, many
	types of integrators can be found in literature, such as Runge-Kutta, Euler, Taylor type integrators and many others. 
	Despite their general applicability, the main problems associated with
	the use of these integrators are that they are often inefficient, and
	prone to instability.
	In \cite{palli} a model of DLO is implemented using a
	multivariate dynamic spline. The integration process is achieved using
	the traditional type Runge-Kutta method. This integration process is
	very time consuming and provide unstable results with many stuck
	situations. Also, it is not practical for a long-time
	experiment. Therefore, it is important to find another integration
	method to reduce simulation time.
	
	The dynamics of a mechanical system, such as a DLO, can be easily
	represented by means of Hamiltonian equations. It is also known that a
	symplectic transformation is the solution of a Hamiltonian system
	\cite{kinoshita}. 
	The numerical solutions obtained by several
	numerical techniques, such as the Runge-Kutta method and the primitive
	Euler scheme, are not energy-preserving map, resulting in spuriously damped or
	excited behaviors.
	
	On the other hand, the symplectic integrator proposed in \cite{feng} exploits the energy-preserving symplectic transformation to avoid spuriously damped or excited solutions. The symplectic integrator scheme has been widely applied to
	the calculations of the long-term evolution of chaotic Hamiltonian
		systems \cite{hu7}, and \cite{hu8}. In the actions of a Hamiltonian system, the symplectic
	integrator does not generate a secular truncation error. The number of
	force function evaluations of the fourth-order symplectic integrator is smaller than the ones of the Runge-Kutta integrator of the same order. The energy conservation and the long-time stability for the symplectic scheme are investigated and verified in \cite{feng}, \cite{hu1}, \cite{hu4}, and \cite{hu9}.
	There are many merits for
	symplectic integrator which are discussed and outlined in deeper
	detail in \cite{forest}, \cite{ruth}, and \cite{neri}. For these
	advantages, the use of the symplectic integrator is investigated in
	this paper for DLO simulation instead of the conventional
	integrators. To this end, a comparison between the results obtained
	from this symplectic integrator and other types of integrators will be
	executed.
	
	
	The main objective of this paper is to compare the properties and results obtained by the symplectic integrator with the ones obtained with other numerical methods. This activity is justified by the fact that, at the best of our knowledge, no data are available in literature about the performances of numerical integration algorithms for DLOs.
	For the DLO modeling, multivariate dynamic splines are used in this paper. A comparison between
	the symplectic, Runge-Kutta, and Zhai methods is performed in
	two cases. In the first case, the effect of gravitational and
	internal forces only are considered, while in the second case the application of an
	external force in addition to the previously mentioned
	forces is considered. 
	
	The remainder of this article is arranged as follows. Sec. 2 outlines
	the key features of the mathematical model of the DLO. The conversion
	from Lagrangian into Hamiltonian and the symplectic integrator are
	discussed in Sec 3. Moreover, Sec. 4 introduces the preliminary
	simulation results and comparisons. Finally, conclusions and future
	work are draft out in Sec. 5.

	\section{Dynamic Model of the DLO }
	A third-order spline basis can effectively represent the shape of the
	DLO. It is a function of a free coordinate $u$. This coordinate $u$
	represents the position on the DLO which is equal to zero at the first
	endpoint. Also, $u$ equals $L$ at the opposite endpoint, provided that
	$L$ is the length of the DLO. This can be written mathematically as:
	\begin{equation}
		q(u)=\sum_{i=1}^{n_{u}} b_{i}(u) q_{i} ,
	\end{equation}
	where $q(u)=(x(u), y(u), z(u), \theta(u))=(r(u), \theta(u))$ is the
	fourth-dimensional configuration functional space of the DLO. It
	includes the three linear coordinates $x, y, z$ of the DLO position at
	point $u$, and the DLO’s axial twisting $\theta$. Also, $b_i (u)$ is
	the $i^{th}$ spline polynomial basis employed to
	describe the shape of the DLO. Moreover, $q_i$ are properly chosen
	coefficients, typically called 
	control points, used to correctly interpolate the shape of the DLO
	through the $b_i(u)$ basis functions. $n_u$ is the number of control points.
	
	For a variety of reasons, this DLO mathematical model is highly successful. Firstly, the spatial derivatives calculation is straightforward, i.e.
	\begin{equation}
		q^{(\mathrm{j})}(u)=\sum_{i=1}^{n_{u}} b_{i}^{(\mathrm{j})}(u) q_{i} ,
	\end{equation}
	where $	q^{(\mathrm{j})}$ is the $j^{th}$ derivative of $q$. In fact, it can be defined by the same coefficients and easy-to-compute derivatives of $b_i(u)$. Second, the spline basis proprieties guarantee that the DLO model curvature, which represents the DLO’s physical behavior \cite{boor}, is minimized. Third, this model enables the representation of a generalized nonlinear function with  smoothness characteristics as a linear combination of the nonlinear function basis $b_i(u)$, which relies on the free variable $u$ only, by the linear coefficients $q_i$.
	
	Referring to the system’s Lagrange equations, the DLO’s dynamic model can be expressed as a function of the control points $q_i$ as:
	\begin{equation}
		\label{eq:lagrange}
		\frac{d}{d t}\left(\frac{\partial K}{\partial \dot{q}_{i}}\right)=F_{i}-\frac{\partial U}{\partial q_{i}}, \quad \quad \quad  \; i=1, 2, \ldots, n_{u},
	\end{equation} 
	where $K$ is the total kinetic energy, $F_i$ is the external force that is applied to the $i^{th}$ control point, and $U$ is the total potential energy acting on the DLO. This potential energy is generated due to the influence of gravity, stretching, torsion, and bending effects on the DLO.
	
	\subsection{Kinetic Energy of the DLO }
	
	Owing to the translational motion of the control points and the rotational motion of the cross sections, the total kinetic energy is generated. In order to express this total kinetic energy, a function of the control points $q_i$ can be used as : 
	\begin{equation}
		K=\frac{1}{2} \int_{0}^{L} \frac{d q^{T}}{d t} J \frac{d q}{d t} \; ds, \quad J=\left[\begin{array}{cccc}
			\mu & 0 & 0 & 0 \\
			0 & \mu & 0 & 0 \\
			0 & 0 & \mu & 0 \\
			0 & 0 & 0 & I
		\end{array}\right] ,
	\end{equation}
	where $J$ is the DLO’s generalized density matrix,  $ds$ is the displacement element and can be computed from $d s=\left\|r^{\prime}(u)\right\| d u$, $\mu$ denotes the linear density, and $I$ denotes the polar moment of inertia. 
	
	According to the procedure explained in \cite{nocent}, we can drive:
	\begin{equation}
		\label{eq:kinetic_energy}
		\frac{d}{d t}\left(\frac{\partial K}{\partial \dot{q}_{i}}\right)=\sum_{j=1}^{n_{u}} M_{i j} \ddot{q}_{j} .
	\end{equation}
	
	Then, by expanding this definition to the overall system, the total inertial forces of the DLO can be written as $M \ddot{q}$, in which $M$ denotes the inertia matrix of the DLO and $\ddot{q}$ is a vector that represents the control point accelerations.

	\subsection{Potential Energy of the DLO }
	
	The total potential energy $U$ is made up of the gravitational effect and the strain energy of the DLO due to stretching, torsion and bending. Although the derived energy of gravity is quite simple, the concept of strain energy plays a crucial role in the modeling of DLO. By introducing the strain vector $\epsilon=\left[\epsilon_{s}, \epsilon_{t}, \epsilon_{b}\right]$  that includes $\epsilon_s$ to represent the stretching term, $\epsilon_t$  to denotes the torsional term, and $\epsilon_b$ to denotes the bending term such that:
	
	\begin{equation}
		\begin{aligned}
			\epsilon_{s} &=1-\left\|r^{\prime}\right\|, \quad \epsilon_{t}=\theta^{\prime}-\gamma, \quad \epsilon_{b}=\frac{\|\mathcal{C}\|}{\left\|r^{\prime}\right\|^{3}} , \\
			\mathcal{C} &=r^{\prime} \times r^{\prime \prime}, \quad \gamma=\frac{\mathcal{C}^{T} r^{\prime \prime \prime}}{\|\mathcal{C}\|^{2}} ,
		\end{aligned}
	\end{equation}
	where $r$ and $\theta$ are the strain’s linear and the torsional component, respectively. The strain energy can therefore be written as:
	
	\begin{equation}
		U=\frac{1}{2} \int_{0}^{L}\left(\epsilon-\epsilon_{0}\right)^{T} H\left(\epsilon-\epsilon_{0}\right) ds=\frac{1}{2} \int_{0}^{L} \epsilon_{e}^{T} H \epsilon_{e} \; ds ,
	\end{equation}
	where 
	$$
	H=\frac{D^{2} \pi}{4}\left[\begin{array}{ccc}E & 0 & 0 \\ 0 & \frac{G D^{2}}{8} & 0 \\ 0 & 0 & \frac{E D^{2}}{16}\end{array}\right]
	$$
	$\epsilon_0$ is the DLO’s plastic strain, which enables to consider the plasticity of the material. $\epsilon_e$ is called the residual strain that equals $\epsilon - \epsilon_0$. Also, $H$ is the element stiffness matrix, $D$ is the cross-section diameter of the DLO. Moreover, $E$ and $G$ denote the Young’s modulus and the shear modulus of the DLO’s material, respectively.
	
	It is possible to write the right-side term of Eq.~\eqref{eq:lagrange} as:
	
	\begin{equation}
		\label{eq:elastic_forces}
		P_{i}=-\frac{\partial U}{\partial q_{i}}=-\frac{1}{2} \int_{0}^{L} \frac{\partial \epsilon_{e}^{T} H \epsilon_{e}}{\partial q_{i}} d s=-\int_{0}^{L} \frac{\partial \epsilon^{T}}{\partial q_{i}} H \epsilon_{e} \; ds ,
	\end{equation}
	where $P_i$  represents the elastic forces because of the DLO deflection. 
	%
	%

	\subsection{Dynamic Model of the DLO}
	
	To write the overall dynamic model of the DLO, the Eqs.~\eqref{eq:lagrange},~\eqref{eq:kinetic_energy}, and~\eqref{eq:elastic_forces} are extended to the whole control points
	
	\begin{equation}
		\label{eq:dlo_dynamics}
		M \ddot{q} = F + P
	\end{equation}
	where $F$ refers to the vector that contains all external forces, including gravity, while $P$ represents the vector that contains all elastic forces.
	
	At every simulation step, the system can be solved by utilizing a simple LU decomposition. Moreover, at each time step, the accelerations are integrated to get the control points velocities and positions. Several integration methods can be used, but some problems may arise, including numerical instability,  large time required for the integration process and the unsuitability for long-time predictions. These problems can be mitigated by using the symplectic integrator, as will be shown in the following. The next section will discuss this symplectic integrator and its requirements in more details.

	\section{Symplectic Integrator}
	
	To solve numerically the differential equations, we decided to use the symplectic integrator because of its advantages with respect to other methods as will be shown in the following. These subsections will discuss in detail calculating the Hamiltonian from Lagrangian and the construction of the symplectic integrator.
	
	\subsection{Conversion from a Lagrangian to a Hamiltonian }
	
	Symplectic integrator is widely used in nonlinear dynamics. It is designed as the numerical solution for Hamilton's equations, which provided as:
	
	\begin{equation}
		\dot{p}=-\frac{\partial H_m}{\partial q} \quad \text { and } \quad \dot{q}=\frac{\partial H_m}{\partial p} ,
	\end{equation}
	where $p$ denotes the momentum coordinates, $q$ refers to the position coordinates, and $H_m$ is the Hamiltonian which can be found from:
	
	\begin{equation}
		H_m(p, q) = K(p) + U(q) ,
	\end{equation}
	where $K$ and $U$ denote the kinetic and potential energy, respectively. The set of position and momentum coordinates $(q, p)$ are named canonical coordinates. In our case, the first step to use the symplectic integrator is to make a conversion from Lagrangian into Hamiltonian \cite{hamill}. This conversion is achieved according to the following procedure.

	Provided a Lagrangian $L_g$ as a function of the generalized coordinates $q_i$ and generalized velocities $\dot q_i$, where $L_g \left(q_{i}, \dot{q}_{i}\right)=K\left(\dot{q}_{i}\right) - U\left(q_{i}\right)$, the Hamiltonian can be calculated according to the following steps:
	\begin{enumerate}
		\item{By differentiating the Lagrangian $L_g$ with respect to the generalized velocities $\dot q_i$, the momenta $p_i$  are determined:}
		
		%
		
		\begin{equation}
			p_{i}\left(q_{i}, \dot{q}_{i}\right)=\frac{\partial L_g}{\partial \dot{q}_{i}}=\frac{\partial \mathrm{K}}{\partial \dot{q}_{i}} . 
		\end{equation}
		
		
		\item{By inverting the expressions in the former step, the velocities $\dot q_i$ are formulated in terms of the momenta $p_i$.}
		
		\begin{equation}
			p_{i}=J q_i^{\cdot} \quad \text { so, } \dot{q_i}=J^{-1} p_{i} ,
		\end{equation}
		where
		$	J^{-1}=\left[\begin{array}{cccc}
			1 / \mu & 0 & 0 & 0 \\
			0 & 1 / \mu & 0 & 0 \\
			0 & 0 & 1 / \mu & 0 \\
			0 & 0 & 0 & 1 / I
		\end{array}\right] .
		$
		
		\item{Using the Lagrangian relation ( $L_g = K - U$ ), we can conclude that: }
		\begin{equation}
			L_g=\frac{1}{2} \int_{0}^{L} \dot q^{T} J \dot{q} \; ds - U .
		\end{equation}
		
		The velocities $\dot q_i$ are then substituted from Equ. (13),
		
		\begin{equation}
			L_g=\frac{1}{2} \int_{0}^{L} p^{T} J^{-1} p \; ds - U .\\
		\end{equation}
		
		\item{The Hamiltonian is determined by employing the typical
			definition of $H_m$ as the Legendre transformation of $L_g$:}
		
		\begin{equation}
			H_m=\sum \dot{q}_{i} \frac{\partial L_g}{\partial \dot{q}_{i}} - L_g = \sum \dot{q}_{i} \, p_{i} - L_g . \\
		\end{equation}
		
		Substitution for $\dot q_i$ from Equ. (13) and the Lagrangian $L_g$ from Equ. (15) into Equ. (16), will lead to: 
		\begin{equation}
			H_m=\frac{1}{2} \int_{0}^{L} p^{T} J^{-1} p \; ds + U .
		\end{equation}
		
		The last equation is the require one as it is equivalent to the Hamiltonian equation that stated in Equ. (11).
	\end{enumerate}
	
	\subsection{Symplectic Integrator of the fourth-order }
	
	Forest \cite{forest} and Neri \cite{neri} introduced the generalized form of the fourth-order symplectic integrator as:
	
	\begin{equation}
		\begin{array}{ll}
			q_{1}=q_{0}+c_{1} \tau \frac{\partial K}{\partial p}\left(p_{0}\right), &\quad p_{1}=p_{0}-d_{1} \tau \frac{\partial U}{\partial q}\left(q_{1}\right) ,\\
			\\
			q_{2}=q_{1}+c_{2} \tau \frac{\partial K}{\partial p}\left(p_{1}\right), &\quad p_{2}=p_{1}-d_{2} \tau \frac{\partial U}{\partial q}\left(q_{2}\right) ,\\
			\\
			q_{3}=q_{2}+c_{3} \tau \frac{\partial K}{\partial p}\left(p_{2}\right), &\quad p_{3}=p_{2}-d_{3} \tau \frac{\partial U}{\partial q}\left(q_{3}\right) ,\\
			\\
			q_{4}=q_{3}+c_{4} \tau \frac{\partial K}{\partial p}\left(p_{3}\right), &\quad p_{4}=p_{3}-d_{4} \tau \frac{\partial U}{\partial q}\left(q_{4}\right) ,
			\\
		\end{array}
	\end{equation}
	where $\tau$  is  the time step-size, $q_0$ and $p_0$ are the initial values, and $q_4$ and $p_4$ are the numerical solution after $\tau$. Moreover, $c_i$  and $d_i$ are numerical coefficients which can be determined uniquely from:
	
	\begin{equation}
		\begin{array}{c}
			c_{1}=c_{4}=\frac{1}{2(2-\beta)}, \quad c_{2}=c_{3}=\frac{(1-\beta)}{2(2-\beta)} , \\
			\\
			d_{1}=d_{3}=\frac{1}{(2-\beta)}, \quad d_{2}=\frac{-\beta}{(2-\beta)}, \quad d_{4}=0 , \\
			\\
			\text { where } \beta=2^{1 / 3} .
		\end{array}
	\end{equation}
	The reader can refer to the reference \cite{ruth} for the derivation and the values of the numerical coefficients.
	
	The number of force function evaluations
	$\frac{\partial U}{\partial q}$, that is the most time-expensive operation
	in the integration process, is three rather than four since
	$d_4=0$. On the other hand, four force function evaluations are
	required in the conventional fourth-order Runge-Kutta integrator. Thus,
	the CPU time can be decreased by approximately $25$
	percent with the symplectic integrator with respect to the fourth-order Runge-Kutta method.
	
	Other than the energy-preserving property, a remarkable characteristic
	of the symplectic integrator is that the aggregation of the truncation
	error in the total energy has not a secular component and the
	positional errors rise in proportional to the first order of time
	\cite{kinoshita}. On the other hand, conventional integrators generate
	a secular energy error. Moreover, the error due to position discretization 
	increase with the square of time. The symplectic integrator is thus well suited to the long-time study of a dynamic system. Generally, in the actions of a Hamiltonian system, the
	symplectic integrator does not generate a secular truncation error
	\cite{kinoshita}.
	
	
	\section{Simulation Results}

	Numerical simulations have been carried out in MATLAB for the
	assessment of the mathematical framework discussed in the preceding
	sections. The issue that is aimed in this section is determining the
	DLO state evolution, i.e. the control points evolution over time, while
	influenced by a internal elastic
	and inertial forces as well as external forces like gravitational force or
	contacts forces generated by obstacles. 
	In the case of contacts, we can reshape the system as a constrained dynamic system in which the Lagrange approach for constraints is adopted. Just add some constraints to the solution of the dynamic system. This will not change the problem, but just increase the number of dynamic equations that existed in the model. In \cite{theetten}, the solution of dynamic equations representing the DLO dynamics including contacts is reported. The same approach can be applied also to the symplectic integrator by  changing the dynamic equations accordingly.
	The simulations have been performed using the dynamic model
	in Eq.~\eqref{eq:dlo_dynamics}. It is worth mentioning that all simulations of this work
	have been performed on Ubuntu 18.04.5 LTS operating system, processor
	intel core i5-3210M CPU@2.50GHz x 4, RAM 8GiB.



	%

\begin{figure*}
			\centering
		\includegraphics[height=6cm, width=0.98\textwidth]{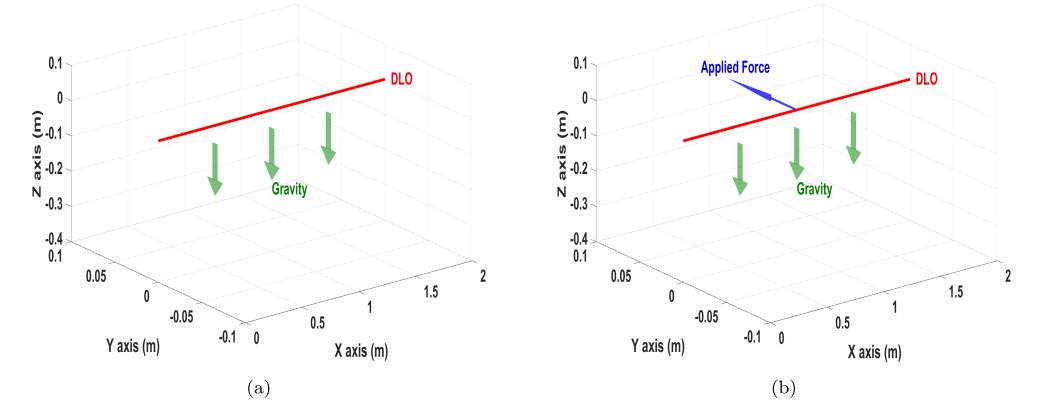}  
		\caption{Schematic diagram of the DLO in which it is affected by: a) gravitational force, and b) gravitational force and the DLO's center is affected by an external force in the Y direction.}
\label{scheme}                
	\end{figure*}

	In our simulations, DLO's material is assumed to be aluminum, and its
	length $L$ is 2 meters. The DLO has a circular cross-section with a diameter D that equals 2 millimeters. The number of control points is selected as
	$n_u$ = 9. Also, the number of sample points along the DLO is chosen
	to be $n_s$=101. These values are chosen to be sufficiently low but
	provide a good interpolation capability. 
	During these simulations, the
	DLO is supposed to be placed initially on a straight position along
	the $x$-axis, and the two extreme endpoints are constrained to move
	along the $x$-axis and attached by springs with stiffness $(K_x=10 \; kN/m)$
	to their initial position. In other words, one endpoint of the DLO is
	located at $(0, 0, 0)$, while the opposite endpoint is located at
	$(L, 0, 0)$.
	Two simulation scenarios will be used in this section. In the first scenario, the
	DLO is affected by the internal forces and the gravitational force only as shown in Fig. \ref*{scheme}a.
	While in the second scenario, the DLO is affected by the previously mentioned forces in addition to an external sinusoidal force which is applied to the DLO center in the Y-direction as illustrated in Fig.~\ref{scheme}b.
	

\begin{figure}
		\includegraphics[height=6cm, width=0.48\textwidth]{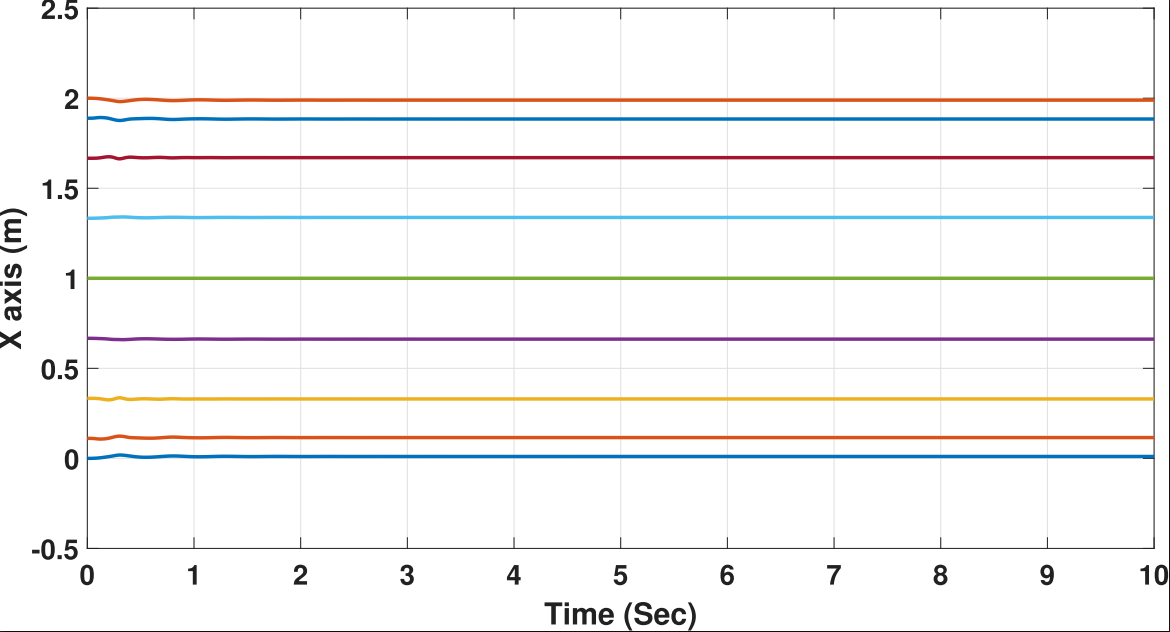}  
		\caption{X component of each control point of the DLO.}        
		\label{xnoforce}                     
	\end{figure}

\begin{figure}
		\includegraphics[height=6cm, width=0.48\textwidth]{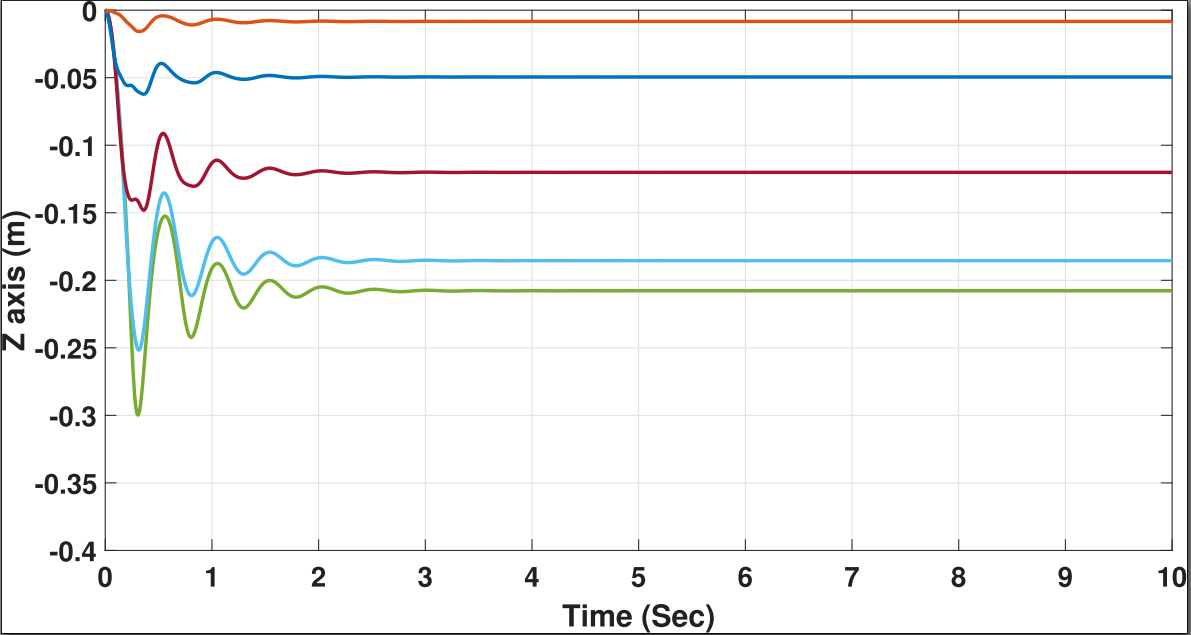}  
		\caption{Z component of each control point of the DLO.}        
		\label{znoforce}                     
	\end{figure}

\begin{figure*}
		\centering
		\includegraphics[height=8cm, width=0.98\textwidth]{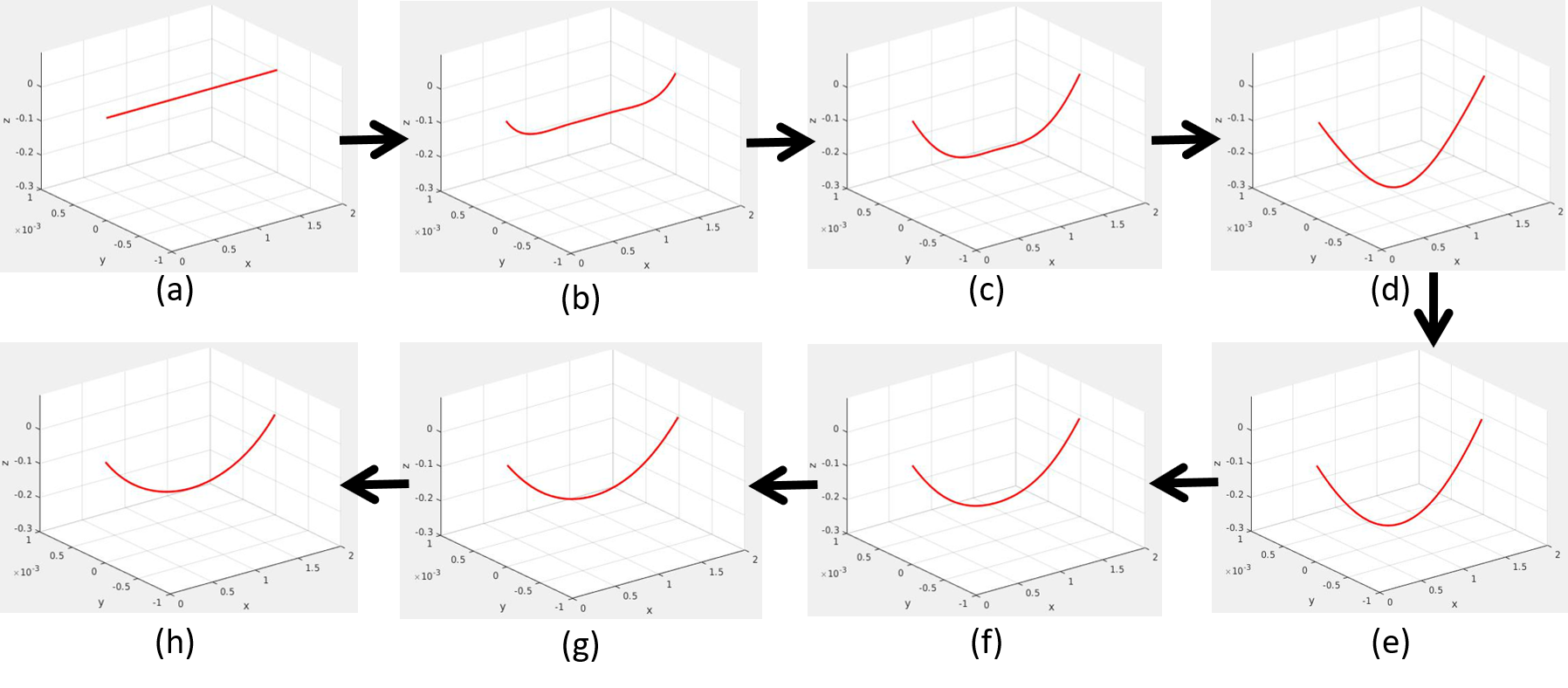}  
		\caption{Sequence of trajectories to show the DLO motion starting from the initial position according to the first scenario.}        
		\label{video1images}                     
	\end{figure*}

	Fig.~\ref{xnoforce}, and \ref{znoforce} illustrate the
	computed solutions accomplished utilizing the spline-based model using
	symplectic integrator where the DLO is affected by the gravity and the
	internal forces only without any other external
	forces. Fig.~\ref{xnoforce} presents the $x$-component of each control
	point, while Fig.~\ref{znoforce} shows the $z$-component of each
	control point. This simulation is achieved for 10 seconds using 2 milliseconds
	as a time step-size $\tau$. This time step-size is the largest value that preserves the stability of the integration method.
	It is worth mentioning that, during the
	whole period, the $y$ component for each control point is equal to zero
	as the DLO is positioned along the $x$-axis and the external force is
	not applied yet. In Fig.~\ref{znoforce}, due to the symmetry of the
	DLO, each two control points’ trajectories are located above each
	other in one curve except the middle point. A video that shows this
	numerical simulation exists in the link
	\footnote{\url{https://www.dropbox.com/s/4tb3ohyod5v71yj/videoforspline.avi?dl=0}}.
	Fig.~\ref{video1images} shows the sequence of trajectories that illustrate the DLO motion starting from the initial position
	according to the first scenario, i.e. the DLO is affected by the internal forces and the gravitational force only without any other external forces.

\begin{figure}
		\includegraphics[height=6cm, width=0.48\textwidth]{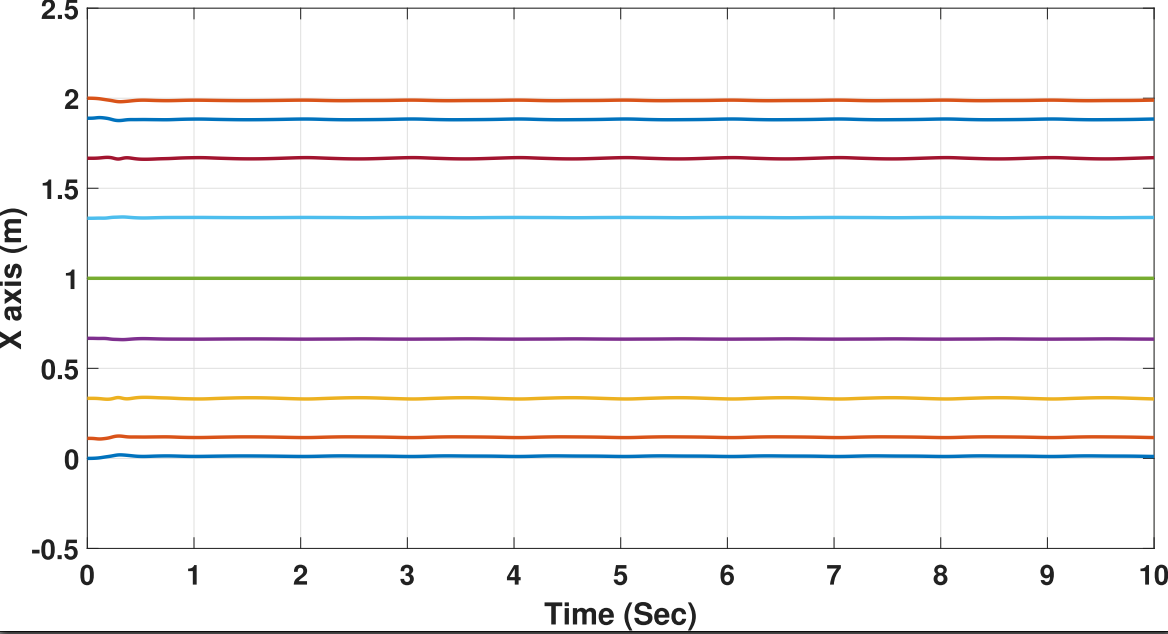}  
		\caption{X component of each control point of the DLO in which its center is affected by external sinusoidal force.}        
		\label{xforce}                     
	\end{figure}

\begin{figure}
		\includegraphics[height=6cm, width=0.48\textwidth]{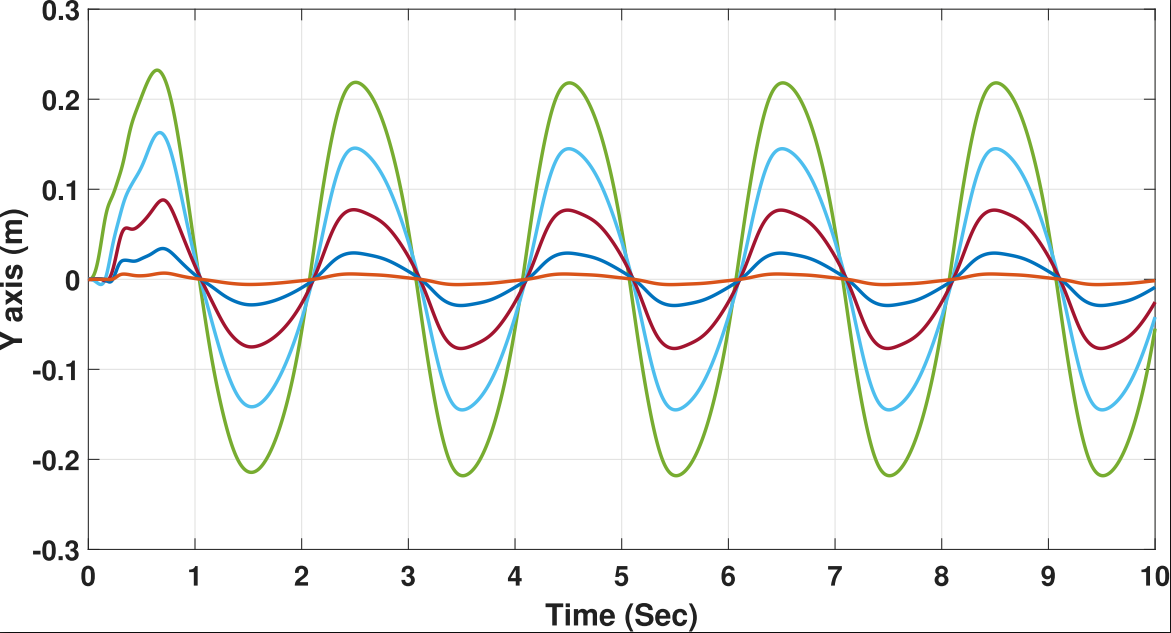}  
		\caption{Y component of each control point of the DLO in which its center is affected by external sinusoidal force.}        
		\label{yforce}                     
	\end{figure}

\begin{figure}
		\includegraphics[height=6cm, width=0.48\textwidth]{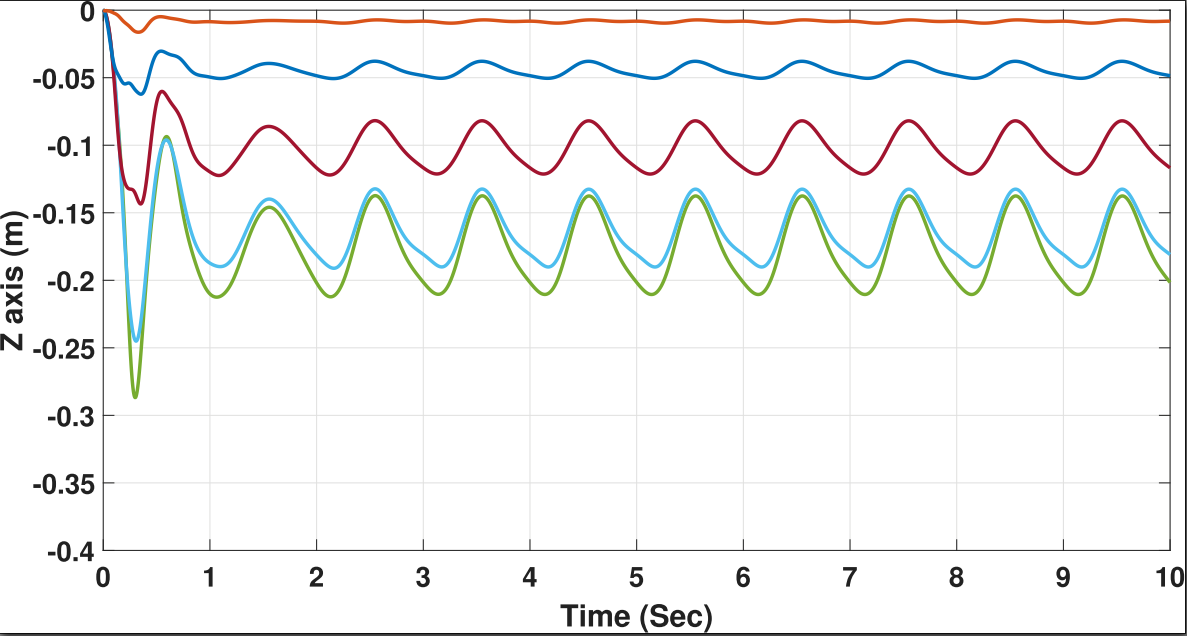}  
		\caption{Z component of each control point of the DLO in which its center is affected by external sinusoidal force.}        
		\label{zforce}                     
	\end{figure}

\begin{figure*}
		\centering
		\includegraphics[height=8cm, width=0.98\textwidth]{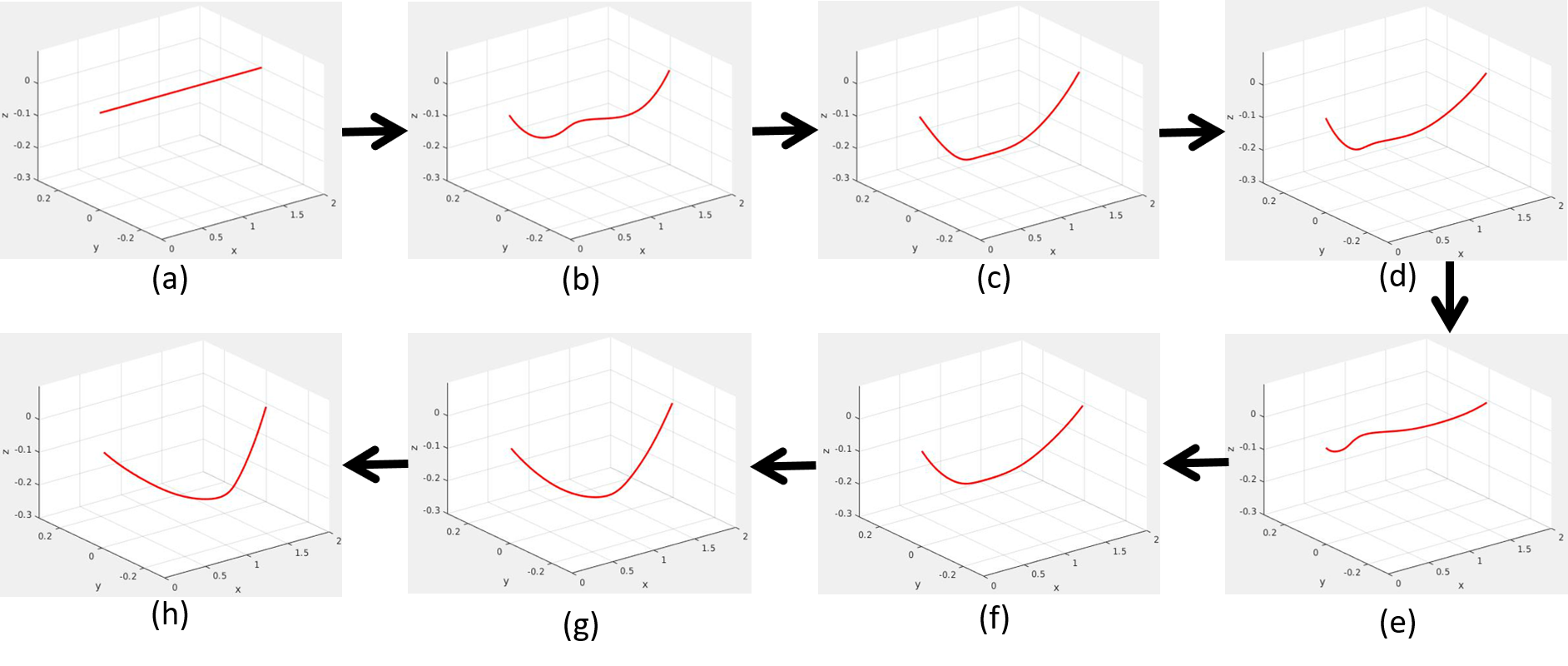}  
		\caption{Sequence of trajectories to show the DLO motion starting from the initial position according to the second scenario.}        
		\label{video2images}                     
	\end{figure*}

	The simulation is performed again in the case of applying an external
	force on a DLO point. We assumed that the external force takes the
	shape of a sinusoidal trajectory. This external force is applied to the
	center of the DLO in the Y-axis direction. Fig.~\ref{xforce},
	\ref{yforce}, and \ref{zforce} show the $x$, $y$, and $z$ components,
	respectively. The solution discussed here assumes the knowledge of the
	external forces $F$ (the input). The computation will give the control
	points $q$ (the output) that describe the DLO motion during the
	simulation interval. A video that illustrates this numerical
	simulation, in case of applying an external force, is found in the
	link
	\footnote{\url{https://www.dropbox.com/s/b1tc8v1k627o33r/videoforspline.avi?dl=0}}. These
	results prove the ability of the spline-based dynamic model and the
	symplectic integrator to represent the state evolution of the DLO
	accurately.
	Fig.~\ref{video2images} shows the sequence of trajectories that illustrate the DLO motion starting from the initial position
	according to the second scenario, i.e. the DLO is affected by the internal forces, the gravitational force, and an external sinusoidal force applied to the DLO center in the Y-direction.
	
	%
	%

	The DLO model simulation time of the symplectic integrator
	is compared with the ones of both the classical Runge-Kutta integrator and Zhai integrator. Zhai integrator is a new simple fast explicit time integration method, the reader can refer to \cite{zhai} for the related details and
	advantages of Zhai integrator. In MATLAB/Simulink, the three
	integrators are implemented by using the same spline-based model
	defined in Sec. 2. In the case the DLO is affected by 
	internal forces and gravity only, Table \ref{tab:matlabnoforce} reports the comparison between the three integrators’ simulation time in seconds for a certain
	period (10 sec) with 2 milliseconds
		as a time step-size $\tau$. The results show that the symplectic integrator is
	faster than both the Runge-Kutta and Zhai
	integration methods.
	It is worth mentioning that the Zhai method needs optimization for two parameters \cite{zhai} . 
		However, we made a manual tunning to these parameters. This may be the reason for
		the notable slowness of Zhai compared to the Runge-Kutta method. 
	
	Table \ref{tab:matlabforce} presents the comparison between the three integrators’
	computation time in seconds for a simulation of 10 seconds in the case
	where the DLO is affected by the external force in addition to the
	previously mentioned forces. The results show that the symplectic
	integrator is able to solve the problem while, on the other hand, the  Runge-Kutta and Zhai integrators have been stuck and could not compute any solution. It is worth mentioning that, in this work, we did not investigate the reason why other methods are unstable.
	
	The simulation implemented in MATLAB, using the fastest integrator,
	takes an average time of 470 seconds for the simulation of 10 seconds
	like the one reported in Fig.~\ref{xnoforce} and \ref{znoforce}. This long execution time occurs
	due to two reasons. The first reason is the specifications of the
	processing unit or the PC. The second reason is the implementation on
	MATLAB which is not really efficient. To avoid this reason, another
	simulation is repeated using C++ language. In these simulations and
	in the related comparisons, the Zhai integrator is excluded as it is the slowest and it looks
	not really efficient in solving the DLO dynamics. Hence, the current
	comparison will be focused on the symplectic and the Runge-Kutta
	integration methods and their implementation in C++. Tables~\ref{tab:cppnoforce} and \ref{tab:cppforce}
	introduce the comparison between the two integrators’ execution time
	in the case of internal forces only and sinusoidal external force,
	respectively. From Tables~\ref{tab:cppnoforce} and \ref{tab:cppforce}, it can be concluded that the symplectic integrator gives the fastest execution time. Also, the C++
	execution time is reduced to be one-third of the execution time of
	MATLAB results.
	
	To check the effect of changes in the DLO model granularity, $n_u$ and
	$n_s$ on the execution time, simulations on both MATLAB and C++ are
	performed. Tables \ref{tab:effectmatlabnoforce} and \ref{tab:effectmatlabforce} present MATLAB results of the execution time
	by changing $n_u$ and $n_s$ for both cases of internal forces only and
	sinusoidal external force, respectively. Each cell in the two tables
	introduces the execution time in seconds for a simulation period of 10
	seconds. Likewise, Table \ref{tab:effectcppnoforce}, and \ref{tab:effectcppforce} illustrate C++ results of execution
	time with changing $n_u$ and $n_s$. From these tables, we conclude
	that C++ is faster about three times more than MATLAB
	implementations. So, C++ implementation with symplectic integrator
	will be better for the practical implementation of the simulation system. Moreover,
	reducing $n_u$ and/or $n_s$ will require less execution time, but it
	will reduce interpolation precision in representing the DLO.
	It is worth mentioning that the time step size is chosen to be unique and equal to 0.8 milliseconds for all of these comparisons. As $n_u$ and $n_s$ are increased, more calculations are required and hence the integration time increases. 
	The same step size is selected for all comparisons for fairness.
	
\begin{figure}
		\includegraphics[height=6cm, width=0.48\textwidth]{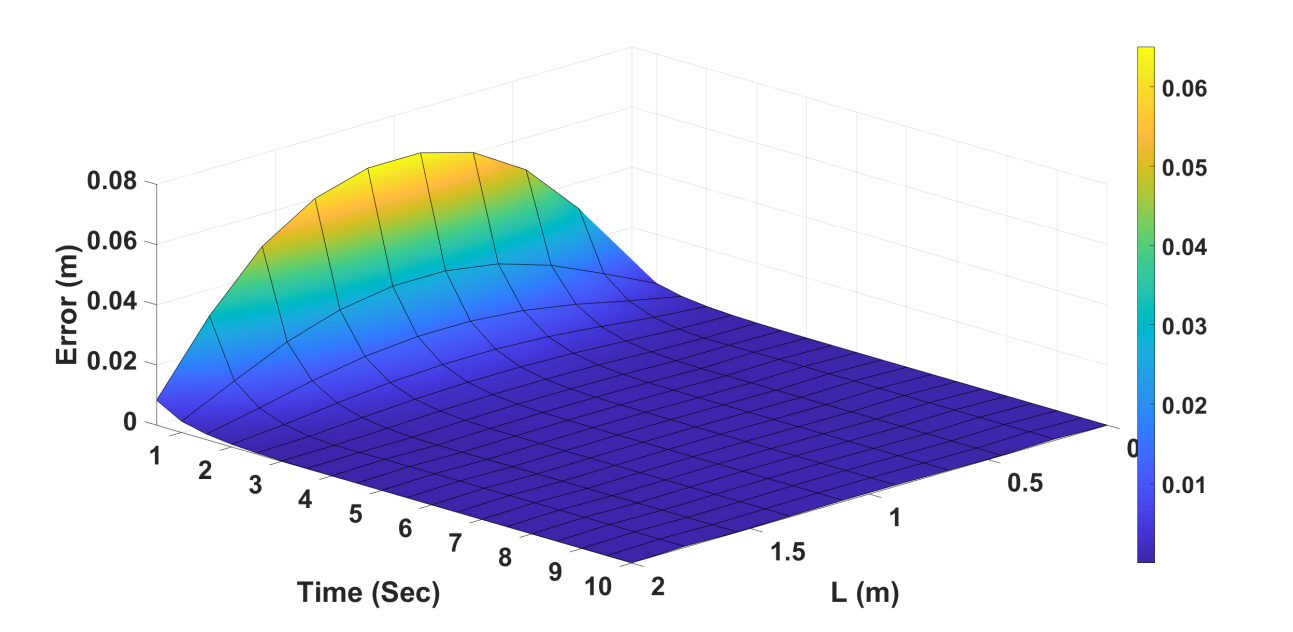}  
		\caption{Evolution of the error over time and DLO length.}        
		\label{f1error}                     
	\end{figure}

\begin{figure}
		\includegraphics[height=6cm, width=0.48\textwidth]{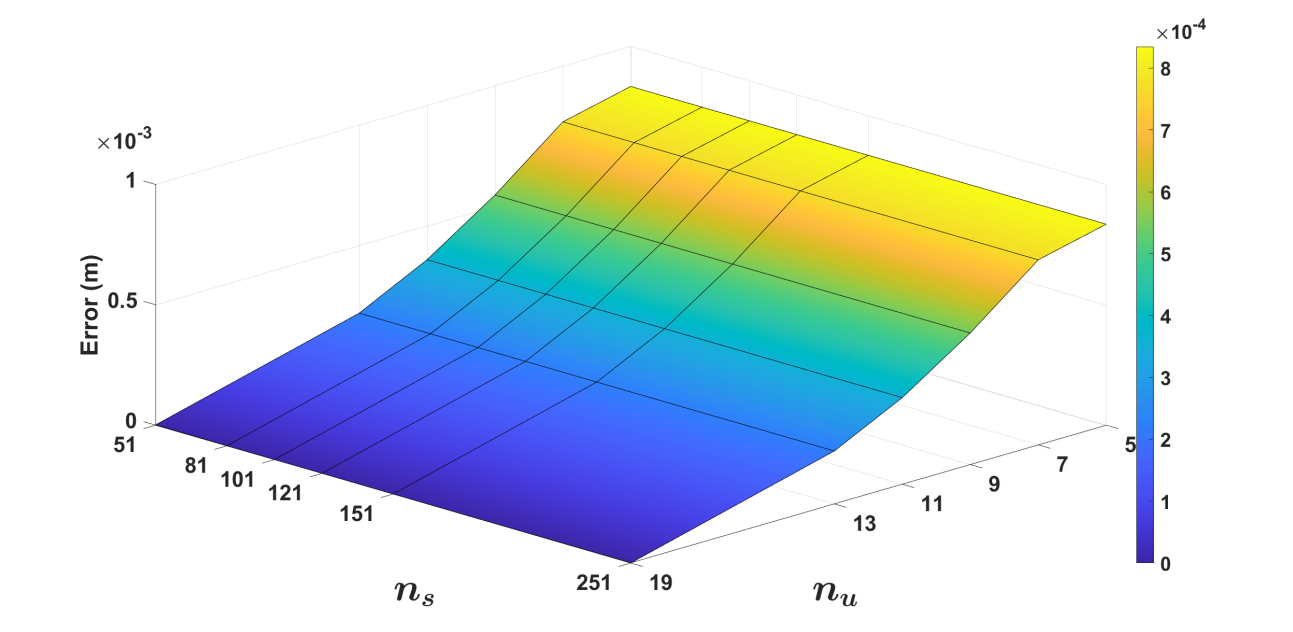}  
		\caption{Evolution of the error over $n_s$ and $n_u$ points.}        
		\label{f2error}                     
	\end{figure}

\begin{figure}
		\includegraphics[height=6cm, width=0.48\textwidth]{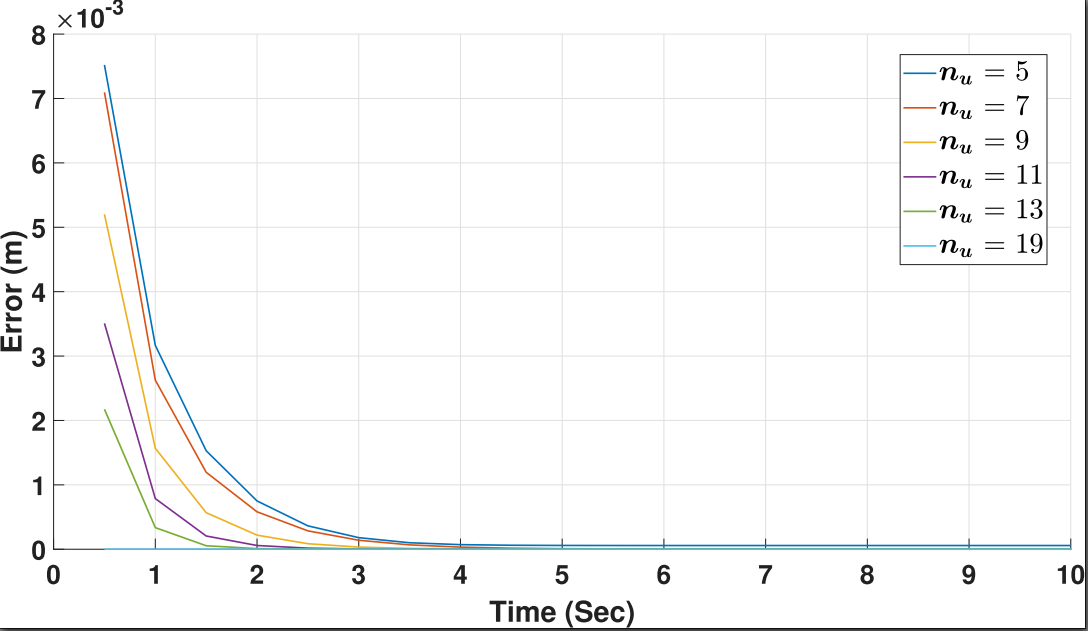}  
		\caption{Evolution of the error over time at different $n_u$ points.}        
		\label{f4error}                     
	\end{figure}

\begin{figure}
		\includegraphics[height=6cm, width=0.48\textwidth]{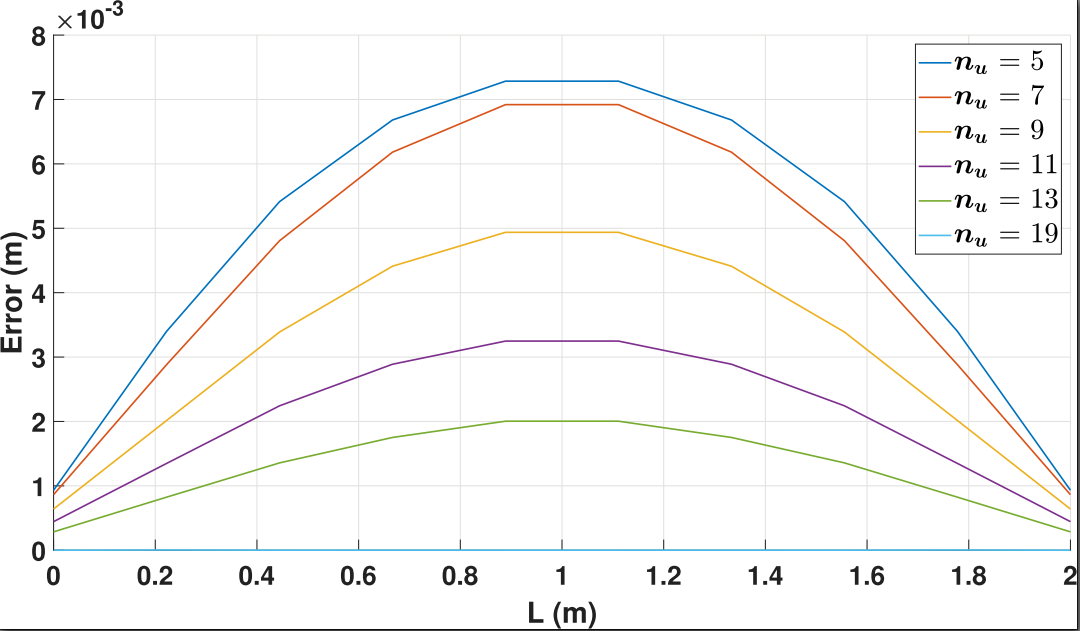}  
		\caption{Evolution of the error over DLO length at different $n_u$ points.}        
		\label{f6error}                     
	\end{figure}

	
	The model with the highest resolution ($n_s = 256$ and $n_u = 19$) is considered as the reference model. Each model resolution considered in this work is compared to the reference model and the error is computed. For each resolution, the DLO position is measured at 10 fixed points along the DLO. Moreover, the error over time is computed by considering 20 equidistant time intervals. Fig.~\ref{f1error} shows the evolution of this error over the time and the DLO length. It is noticed that the maximum error occurred at the center of the DLO and the error decreases over time. The evolution of the error with different values for $n_s$ and $n_u$ is presented in Fig.~\ref{f2error}. As $n_u$ increases, the error decreases and vice versa, while $n_s$ has no noticeable effect on the error. Moreover, as the number of control points $n_u$ increases, the error decreases over time and with respect to the position along the DLO as indicated in Fig.~\ref{f4error} and \ref{f6error}, respectively.


	\begin{table}[h!]
		\caption{MATLAB execution time for solving the DLO model employing different integrators without applying an external force.}   	
		
		\label{tab:matlabnoforce}
		\resizebox{0.46\textwidth}{!}{%
			\begin{tabular}{cccc}
				\hline \hline
				\textbf{Simulation time} & \textbf{Symplectic}  & \textbf{Runge-Kutta} & \textbf{Zhai}    \\ \hline
				0.1                   & 6.66                       & 9.43          & 34.85   \\
				0.2                   & 11.39                       & 16.42         & 67.57   \\ 
				0.3                   & 16.14                       & 22.43         & 96.69   \\ 
				0.4                   & 20.86                       & 28.41         & 120.45  \\ 
				0.5                   & 25.60                       & 34.62         & 145.42  \\ 
				0.6                   & 30.42                       & 41.29         & 166.22  \\ 
				0.7                   & 35.17                       & 48.62         & 186.46  \\ 
				0.8                   & 39.92                       & 56.07         & 206.70  \\ 
				0.9                   & 44.61                       & 62.59         & 226.81  \\ 
				1                     & 49.30                       & 68.89         & 247.39  \\ 
				2                     & 97.18                       & 140.00        & 478.94  \\ 
				3                     & 144.21                      & 210.41        & 685.46  \\
				4                     & 191.20                      & 278.22        & 904.89  \\ 
				5                     & 238.24                      & 347.53        & 1127.37 \\ 
				6                     & 285.31                      & 418.16        & 1348.17 \\ 
				7                     & 332.29                      & 487.72        & 1542.90 \\ 
				8                     & 379.23                      & 555.20        & 1733.71 \\ 
				9                     & 426.17                      & 624.36        & 1970.40 \\ 
				10                    & 473.18                      & 697.04        & 2204.93 \\ \hline \hline
			\end{tabular}%
		}
	\end{table}

	\begin{table}[]
		\caption{MATLAB execution time for solving the DLO model employing different integrators in case of applying an external force.}   	
		\label{tab:matlabforce}
		\resizebox{0.46\textwidth}{!}{%
			\begin{tabular}{cccc}
				\hline \hline
				\textbf{Simulation time} & \textbf{Symplectic} & \textbf{Runge-Kutta} & \textbf{Zhai} \\ \hline
				0.1 &
				6.83 &
				\multirow{19}{*}{\begin{tabular}[c]{@{}c@{}} Unstable  \end{tabular}} &
				\multirow{19}{*}{\begin{tabular}[c]{@{}c@{}} Unstable  \end{tabular}} \\ 
				0.2                   & 11.75                       &               &      \\ 
				0.3                   & 16.70                       &               &      \\ 
				0.4                   & 21.64                       &               &      \\ 
				0.5                   & 26.60                       &               &      \\ 
				0.6                   & 31.56                       &               &      \\ 
				0.7                   & 36.55                       &               &      \\ 
				0.8                   & 41.48                       &               &      \\ 
				0.9                   & 46.44                       &               &      \\ 
				1                     & 51.40                       &               &      \\ 
				2                     & 101.14                      &               &      \\ 
				3                     & 150.73                      &               &      \\ 
				4                     & 200.30                      &               &      \\ 
				5                     & 249.89                      &               &      \\ 
				6                     & 298.60                      &               &      \\ 
				7                     & 346.63                      &               &      \\ 
				8                     & 394.63                      &               &      \\ 
				9                     & 442.50                      &               &      \\ 
				10                    & 490.59                      &               &      \\ \hline \hline
			\end{tabular}%
		}
	\end{table}

	\begin{table}[]
		\caption{C++ execution time for solving the DLO model employing different integrators without applying an external force.}   	
		
		\label{tab:cppnoforce}
		\resizebox{0.46\textwidth}{!}{%
			\begin{tabular}{ccc}
				\hline \hline
				\textbf{Simulation time} & \textbf{Symplectic} & \textbf{Runge-Kutta}    \\ \hline
				0.1                   & 2.04                        & 5.88   \\ 
				0.2                   & 3.85                        & 10.44  \\ 
				0.3                   & 5.65                        & 13.90  \\ 
				0.4                   & 7.55                        & 16.75  \\ 
				0.5                   & 9.38                        & 19.93  \\ 
				0.6                   & 11.17                       & 23.09  \\ 
				0.7                   & 12.95                       & 26.43  \\ 
				0.8                   & 14.73                       & 29.82  \\ 
				0.9                   & 16.56                       & 33.00  \\ 
				1                     & 18.35                       & 36.39  \\ 
				2                     & 36.57                       & 69.57  \\ 
				3                     & 54.67                       & 102.85 \\ 
				4                     & 72.88                       & 136.00 \\ 
				5                     & 90.99                       & 169.06 \\ 
				6                     & 109.11                      & 201.95 \\ 
				7                     & 127.30                      & 235.07 \\ 
				8                     & 145.86                      & 268.21 \\
				9                     & 164.16                      & 301.24 \\ 
				10                    & 182.46                      & 334.40 \\ \hline \hline
			\end{tabular}%
		}
		
	\end{table}

	\begin{table}[]
		\caption{C++ execution time for solving the DLO model employing different integrators in case of applying an external force.}   	
		
		\label{tab:cppforce}
		\resizebox{0.46\textwidth}{!}{%
			\begin{tabular}{ccc}
				\hline \hline
				\textbf{Simulation time} & \textbf{Symplectic} & \textbf{Runge-Kutta }  \\ \hline
				0.1 & 1.98   & 5.45   \\ 
				0.2 & 3.98   & 9.85  \\ 
				0.3 & 5.92   & 13.63  \\ 
				0.4 & 7.93   & 16.88  \\ 
				0.5 & 9.99   & 20.12  \\ 
				0.6 & 11.91  & 23.42  \\ 
				0.7 & 13.97  & 26.81  \\
				0.8 & 15.99  & 30.52  \\ 
				0.9 & 17.90  & 34.53  \\ 
				1   & 19.90  & 38.17  \\ 
				2   & 40.10  & 76.95  \\ 
				3   & 60.16  & 115.53 \\ 
				4   & 80.41  & 154.16 \\ 
				5   & 101.47 & 193.16 \\ 
				6   & 123.51 & 231.97 \\ 
				7   & 146.46 & 270.46 \\ 
				8   & 167.34 & 309.58 \\ 
				9   & 189.53 & 349.07 \\ 
				10  & 210.46 & 388.04 \\ \hline\hline
			\end{tabular}%
		}
		
	\end{table}

	\begin{table}[]
		\caption{Effects of changing the model parameters $(n_u \& n_s)$ on MATLAB execution time for solving the DLO model employing the symplectic integrator without applying an external force.}   	
		
		\label{tab:effectmatlabnoforce}
		\resizebox{0.46\textwidth}{!}{%
			\begin{tabular}{ccccccc}
				\hline \hline
				\multirow{2}{*}{\boldmath{\large $n_u$}} & \multicolumn{6}{c}{\boldmath{\large $n_s$}}                                                   \\ \cline{2-7} 
				& \textbf{51} & \textbf{81} & \textbf{101} & \textbf{121} & \textbf{151} & \textbf{251} \\ \hline
				\textbf{5}                     & 488.28      & 688.22      & 825.99       & 960.52       & 1288.93      & 2094.12      \\ 
				\textbf{7}                     & 588.53      & 811.56      & 998.6        & 1201.32      & 1419.81      & 2357.17      \\ 
				\textbf{9}                     & 713.17      & 967.77      & 1299.68      & 1421.18      & 1634.21      & 2787.05      \\ 
				\textbf{11}                    & 770.00      & 1129.01     & 1344.34      & 1691.19      & 2033.16      & 3171.8       \\ 
				\textbf{13}                    & 908.49      & 1332.11     & 1566.67      & 1893.44      & 2304.9       & 3909.13      \\ 
				\textbf{19}                    & 1331.77     & 1714.65     & 2348.26      & 2479.57      & 3104.75      & 5090.28      \\ \hline \hline
			\end{tabular}
		}
		\label{tab:caption}
		
	\end{table}

	\begin{table}[]
		\caption{Effects of changing the model parameters $(n_u \& n_s)$ on MATLAB execution time for solving the DLO model employing the symplectic integrator in case of applying an external force.}   	
		
		\label{tab:effectmatlabforce}
		\resizebox{0.46\textwidth}{!}{%
			\begin{tabular}{ccccccc}
				\hline \hline
				\multirow{2}{*}{\boldmath{\large $n_u$}} & \multicolumn{6}{c}{\boldmath{\large $n_s$}}                                                   \\ \cline{2-7} 
				& \textbf{51} & \textbf{81} & \textbf{101} & \textbf{121} & \textbf{151} & \textbf{251} \\ \hline
				\textbf{5}                     & 489.98      & 677.93      & 888.11       & 1004.88      & 1254.07      & 2015.19      \\ 
				\textbf{7}                     & 538.59      & 881.56      & 1007.06      & 1214.88      & 1554.83      & 2379.88      \\ 
				\textbf{9}                     & 643.93      & 970.64      & 1124.15      & 1405.01      & 1675.14      & 2970.31      \\
				\textbf{11}                    & 717.9       & 1101.62     & 1385.67      & 1692.81      & 2006.02      & 3140.75      \\ 
				\textbf{13}                    & 874.72      & 1291.78     & 1568.77      & 1886.01      & 2313.28      & 3903.67      \\
				\textbf{19}                    & 1386.62     & 1826.59     & 2102.17      & 2667.17      & 2990.85      & 5035.5       \\ \hline \hline
			\end{tabular}
		}
		
	\end{table}

	\begin{table}[]
		\caption{Effects of changing the model parameters $(n_u \& n_s)$ on C++ execution time for solving the DLO model employing the symplectic integrator without applying an external force.}   	
		
		\label{tab:effectcppnoforce}
		\resizebox{0.46\textwidth}{!}{%
			\begin{tabular}{ccccccc}
				\hline \hline
				\multirow{2}{*}{\boldmath{\large $n_u$}} & \multicolumn{6}{c}{\boldmath{\large $n_s$}}                                                   \\ \cline{2-7} 
				& \textbf{51} & \textbf{81} & \textbf{101} & \textbf{121} & \textbf{151} & \textbf{251} \\ \hline
				\textbf{5}                     & 155.28      & 232.5       & 284.52       & 339.25       & 416.16       & 673.25       \\ 
				\textbf{7}                     & 197.56      & 295.17      & 363.86       & 435.05       & 529.7        & 861.73       \\
				\textbf{9}                     & 245.61      & 369.37      & 455.4        & 523.42       & 652.07       & 1065.55      \\ 
				\textbf{11}                    & 286.957     & 436.69      & 534.16       & 623.1        & 765.75       & 1271.19      \\ 
				\textbf{13}                    & 319.786     & 489.52      & 601.8        & 713.24       & 887.17       & 1437.82      \\ 
				\textbf{19}                    & 450.95      & 690.64      & 840.81       & 990.89       & 1236.14      & 2021.42      \\ \hline \hline
			\end{tabular}
		}
		
	\end{table}

	\begin{table}[]
		\caption{Effects of changing the model parameters $(n_u \& n_s)$ on C++ execution time for solving the DLO model employing the symplectic integrator in case of applying an external force.}   	
		
		\label{tab:effectcppforce}
		\resizebox{0.46\textwidth}{!}{%
			\begin{tabular}{ccccccc}
				\hline \hline
				\multirow{2}{*}{\boldmath{\large $n_u$}} & \multicolumn{6}{c}{\boldmath{\large $n_s$}}                                                   \\ \cline{2-7} 
				& \textbf{51} & \textbf{81} & \textbf{101} & \textbf{121} & \textbf{151} & \textbf{251} \\ \hline
				\textbf{5}                     & 154.18      & 231.83      & 283.43       & 335.97       & 414.29       & 680.41       \\ 
				\textbf{7}                     & 195.2       & 295.25      & 361.43       & 427.65       & 530.54       & 863.73       \\ 
				\textbf{9}                     & 238.25      & 358.64      & 442.07       & 521.7        & 645.43       & 1060.26      \\ 
				\textbf{11}                    & 278.72      & 426.39      & 523.94       & 617.38       & 764.78       & 1253.72      \\ 
				\textbf{13}                    & 319.12      & 488.51      & 600.33       & 709.03       & 889.24       & 1436.29      \\
				\textbf{19}                    & 442.82      & 677.01      & 835.82       & 988.3        & 1228.88      & 2023.9       \\ \hline \hline
			\end{tabular}
		}
		
	\end{table}

	\section{Conclusions}
	This paper reports a comparison among integration methods to solve the
	dynamic model of Deformable Linear Objects. The
	adopted DLO model is based on the multivariate dynamic spline and
	employs the symplectic integrator. This symplectic integrator received particular attention due
	to its advantages over alternative methods in solving the dynamic
	equations of Hamiltonian systems. Simulations on MATLAB and using C++
	code are performed to compare the performance of the symplectic
	integrator against the Runge-Kutta and Zhai integrators. The results
	prove that the symplectic integrator is the fastest and the most
	stable integrator among the considered ones. Moreover, the proposed
	comparison shows that C++ is about three times faster than the
	corresponding MATLAB implementation. Hence, a spline-based model with
	a symplectic integrator looks to be the best choice for practical
	implementation. Finally, comparisons are performed to check the
	effects of changing the model granularity. These results show that reducing the
	control points and sample points will reduce the execution time, but it will
	reduce the interpolation precision in representing the DLO behavior.
	
	In future work, practical experiments will be used to validate the
	spline-based model and the results obtained with the symplectic integrator. Moreover, this model will be exploited to optimize the DLO manipulation process. Also, an
	extension to the method will be applied to multi-branch DLOs, like the
	pre-assembled wiring harness utilized in the aerospace and automotive
	industries.
	
\section*{acknowledgment}
	This work was supported by the European Commissions Horizon 2020
	Framework Programme with the project REMODEL - Robotic technologies
	for the manipulation of complex deformable linear objects - under
	grant agreement No 870133.\\
	


\begin{thebibliography}{10}

\bibitem{boor}
Carl De~Boor.
\newblock A practical guide to splines; rev. ed., ser. applied mathematical
  sciences.
\newblock 2001.

\bibitem{feng}
Kang Feng.
\newblock On difference schemes and symplectic geometry.
\newblock In {\em Proceedings of the 5th international symposium on
  differential geometry and differential equations}, 1984.

\bibitem{forest}
Etienne Forest.
\newblock Canonical integrators as tracking codes (or how to integrate
  perturbation theory with tracking).
\newblock Technical report, Lawrence Berkeley Lab., 1987.

\bibitem{ruth}
Etienne Forest and Ronald~D Ruth.
\newblock Fourth-order symplectic integration.
\newblock {\em Physica D: Nonlinear Phenomena}, 43(1):105--117, 1990.

\bibitem{greco}
Leopoldo Greco and Massimo Cuomo.
\newblock B-spline interpolation of kirchhoff-love space rods.
\newblock {\em Computer Methods in Applied Mechanics and Engineering},
  256:251--269, 2013.

\bibitem{hamill}
Patrick Hamill.
\newblock {\em A student's guide to Lagrangians and Hamiltonians}.
\newblock Cambridge University Press, 2014.

\bibitem{hermansson}
Tomas Hermansson, Robert Bohlin, Johan~S Carlson, and Rikard S{\"o}derberg.
\newblock Automatic assembly path planning for wiring harness installations.
\newblock {\em Journal of manufacturing systems}, 32(3):417--422, 2013.

\bibitem{hu8}
Jialin Hong, Chuying Huang, and Xu~Wang.
\newblock Symplectic runge--kutta methods for hamiltonian systems driven by
  gaussian rough paths.
\newblock {\em Applied Numerical Mathematics}, 129:120--136, 2018.

\bibitem{hu9}
Weipeng Hu, Zichen Deng, Songmei Han, and Wenrong Zhang.
\newblock Generalized multi-symplectic integrators for a class of hamiltonian
  nonlinear wave pdes.
\newblock {\em Journal of Computational Physics}, 235:394--406, 2013.

\bibitem{hu2}
Weipeng Hu, Zhen Wang, Yunping Zhao, and Zichen Deng.
\newblock Symmetry breaking of infinite-dimensional dynamic system.
\newblock {\em Applied Mathematics Letters}, 103:106207, 2020.

\bibitem{hu4}
Weipeng Hu, Mengbo Xu, Jiangrui Song, Qiang Gao, and Zichen Deng.
\newblock Coupling dynamic behaviors of flexible stretching hub-beam system.
\newblock {\em Mechanical Systems and Signal Processing}, 151:107389, 2021.

\bibitem{hu3}
Weipeng Hu, Juan Ye, and Zichen Deng.
\newblock Internal resonance of a flexible beam in a spatial tethered system.
\newblock {\em Journal of Sound and Vibration}, 475:115286, 2020.

\bibitem{hu1}
Weipeng Hu, Tingting Yin, Wei Zheng, and Zichen Deng.
\newblock Symplectic analysis on orbit-attitude coupling dynamic problem of
  spatial rigid rod.
\newblock {\em Journal of Vibration and Control}, 26(17-18):1614--1624, 2020.

\bibitem{hu6}
Weipeng Hu, Lingjun Yu, and Zichen Deng.
\newblock Minimum control energy of spatial beam with assumed attitude
  adjustment target.
\newblock {\em Acta Mechanica Solida Sinica}, 33(1):51--60, 2020.

\bibitem{hu5}
Weipeng Hu, Chuanzeng Zhang, and Zichen Deng.
\newblock Vibration and elastic wave propagation in spatial flexible damping
  panel attached to four special springs.
\newblock {\em Communications in Nonlinear Science and Numerical Simulation},
  84:105199, 2020.

\bibitem{huang}
Sandy~H Huang, Jia Pan, George Mulcaire, and Pieter Abbeel.
\newblock Leveraging appearance priors in non-rigid registration, with
  application to manipulation of deformable objects.
\newblock In {\em 2015 IEEE/RSJ International Conference on Intelligent Robots
  and Systems (IROS)}, pages 878--885. IEEE, 2015.

\bibitem{jayender}
Jagadeesan Jayender, Rajnikant~V Patel, and Suwas Nikumb.
\newblock Robot-assisted active catheter insertion: Algorithms and experiments.
\newblock {\em The International Journal of Robotics Research},
  28(9):1101--1117, 2009.

\bibitem{jiang}
Xin Jiang, Kyong-mo Koo, Kohei Kikuchi, Atsushi Konno, and Masaru Uchiyama.
\newblock Robotized assembly of a wire harness in a car production line.
\newblock {\em Advanced Robotics}, 25(3-4):473--489, 2011.

\bibitem{jimenez}
P~Jim{\'e}nez.
\newblock Survey on model-based manipulation planning of deformable objects.
\newblock {\em Robotics and computer-integrated manufacturing}, 28(2):154--163,
  2012.

\bibitem{kinoshita}
Hiroshi Kinoshita, Haruo Yoshida, and Hiroshi Nakai.
\newblock Symplectic integrators and their application to dynamical astronomy.
\newblock {\em Celestial Mechanics and Dynamical Astronomy}, 50:59--71, 1991.

\bibitem{linn}
Joachim Linn and Klaus Dre{\ss}ler.
\newblock Discrete cosserat rod models based on the difference geometry of
  framed curves for interactive simulation of flexible cables.
\newblock In {\em Math for the Digital Factory}, pages 289--319. Springer,
  2017.

\bibitem{tangled}
Wen~Hao Lui and Ashutosh Saxena.
\newblock Tangled: Learning to untangle ropes with rgb-d perception.
\newblock In {\em 2013 IEEE/RSJ International Conference on Intelligent Robots
  and Systems}, pages 837--844. IEEE, 2013.

\bibitem{lvphysically}
Naijing Lv, Jianhua Liu, Xiaoyu Ding, Jiashun Liu, Haili Lin, and Jiangtao Ma.
\newblock Physically based real-time interactive assembly simulation of cable
  harness.
\newblock {\em Journal of Manufacturing Systems}, 43:385--399, 2017.

\bibitem{lv}
Naijing Lv, Jianhua Liu, Huanxiong Xia, Jiangtao Ma, and Xiaodong Yang.
\newblock A review of techniques for modeling flexible cables.
\newblock {\em Computer-Aided Design}, 122:102826, 2020.

\bibitem{masey}
Rene J~Moreno Masey, John~O Gray, Tony~J Dodd, and Darwin~G Caldwell.
\newblock Guidelines for the design of low-cost robots for the food industry.
\newblock {\em Industrial Robot: An International Journal}, 2010.

\bibitem{moll}
Mark Moll and Lydia~E Kavraki.
\newblock Path planning for deformable linear objects.
\newblock {\em IEEE Transactions on Robotics}, 22(4):625--636, 2006.

\bibitem{neri}
Filippo Neri.
\newblock Lie algebras and canonical integration.
\newblock {\em Dept. of Physics, University of Maryland}, 1987.

\bibitem{nocent}
Olivier Nocent and Yannick Remion.
\newblock Continuous deformation energy for dynamic material splines subject to
  finite displacements.
\newblock In {\em Computer Animation and Simulation 2001}, pages 87--97.
  Springer, 2001.

\bibitem{palli}
Gianluca Palli.
\newblock Model-based manipulation of deformable linear objects by multivariate
  dynamic splines.
\newblock In {\em 2020 IEEE Conference on Industrial Cyberphysical Systems
  (ICPS)}, volume~1, pages 520--525. IEEE, 2020.

\bibitem{rambow}
Matthias Rambow, Thomas Schau{\ss}, Martin Buss, and Sandra Hirche.
\newblock Autonomous manipulation of deformable objects based on teleoperated
  demonstrations.
\newblock In {\em 2012 IEEE/RSJ International Conference on Intelligent Robots
  and Systems}, pages 2809--2814. IEEE, 2012.

\bibitem{ramisa}
Arnau Ramisa, Guillem Alenya, Francesc Moreno-Noguer, and Carme Torras.
\newblock Using depth and appearance features for informed robot grasping of
  highly wrinkled clothes.
\newblock In {\em 2012 IEEE International Conference on Robotics and
  Automation}, pages 1703--1708. IEEE, 2012.

\bibitem{sanchez}
Jose Sanchez, Juan-Antonio Corrales, Belhassen-Chedli Bouzgarrou, and Youcef
  Mezouar.
\newblock Robotic manipulation and sensing of deformable objects in domestic
  and industrial applications: a survey.
\newblock {\em The International Journal of Robotics Research}, 37(7):688--716,
  2018.

\bibitem{hu7}
JM9728120655 Sanz-Serna.
\newblock Runge-kutta schemes for hamiltonian systems.
\newblock {\em BIT Numerical Mathematics}, 28(4):877--883, 1988.

\bibitem{servin}
Martin Servin and Claude Lacoursiere.
\newblock Rigid body cable for virtual environments.
\newblock {\em IEEE Transactions on Visualization and Computer Graphics},
  14(4):783--796, 2008.

\bibitem{shah}
Ankit Shah, Lotta Blumberg, and Julie Shah.
\newblock Planning for manipulation of interlinked deformable linear objects
  with applications to aircraft assembly.
\newblock {\em IEEE Transactions on Automation Science and Engineering},
  15(4):1823--1838, 2018.

\bibitem{theetten}
Adrien Theetten, Laurent Grisoni, Claude Andriot, and Brian Barsky.
\newblock Geometrically exact dynamic splines.
\newblock {\em Computer-Aided Design}, 40(1):35--48, 2008.

\bibitem{quasi}
Adrien Theetten, Laurent Grisoni, Christian Duriez, and Xavier Merlhiot.
\newblock Quasi-dynamic splines.
\newblock In {\em Proceedings of the 2007 ACM symposium on Solid and physical
  modeling}, pages 409--414, 2007.

\bibitem{valentini}
Pier~Paolo Valentini and Ettore Pennestr{\`\i}.
\newblock Modeling elastic beams using dynamic splines.
\newblock {\em Multibody system dynamics}, 25(3):271--284, 2011.

\bibitem{wang}
Weifu Wang, Dmitry Berenson, and Devin Balkcom.
\newblock An online method for tight-tolerance insertion tasks for string and
  rope.
\newblock In {\em 2015 IEEE International Conference on Robotics and Automation
  (ICRA)}, pages 2488--2495. IEEE, 2015.

\bibitem{zhai}
Wanming Zhai.
\newblock Numerical method and computer simulation for analysis of
  vehicle--track coupled dynamics.
\newblock In {\em Vehicle--track coupled dynamics}, pages 203--229. Springer,
  2020.

\end{thebibliography}

\end{document}